\documentclass[a4paper]{lipics-v2021}

\usepackage{multicol}
\usepackage{adjustbox}
\usepackage{multirow}
\usepackage{listings}
\usepackage{pifont}
\usepackage{xcolor}

\title{Semi-supervised Learning from Street-View Images and OpenStreetMap for Automatic Building Height Estimation} 

\titlerunning{Automatic Building Height Estimation}

\author{Hao Li}{Technical University of Munich, hao\_bgd.li@tum.de}{}{}{}

\author{Zhendong Yuan}{Utrecht University, z.yuan@uu.nl}{}{}{}

\author{Gabriel Dax}{Technical University of Munich, gabriel.dax@tum.de}{}{}{}

\author{Gefei Kong}{Norwegian University of Science and Technology, gefei.kong@ntnu.no}{}{}{}

\author{Hongchao Fan}{Norwegian University of Science and Technology, hongchao.fan@ntnu.no}{}{}{}

\author{Alexander Zipf}{GIScience Chair, Heidelberg University, zipf@uni-heidelberg.de}{}{}{}

\author{Martin Werner}{Technical University of Munich, martin.werner@tum.de}{}{}{}

\authorrunning{Li, H. et al}

\Copyright{Hao Li, Zhendong Yuan, Gabriel Dax, Gefei Kong, Hongchao Fan, Alexander Zipf, Martin Werner}



\relatedversion{}

\supplement{}

\nolinenumbers 

\hideLIPIcs  

\EventEditors{Roger Beecham, Jed A. Long, Dianna Smith, Qunshan Zhao, and Sarah Wise}
\EventNoEds{5}
\EventLongTitle{12th International Conference on Geographic Information Science (GIScience 2023)}
\EventShortTitle{GIScience 2023}
\EventAcronym{GIScience}
\EventYear{2023}
\EventDate{September 12--15, 2023}
\EventLocation{Leeds, UK}
\EventLogo{}
\SeriesVolume{277}
\ArticleNo{4}
\begin{document}

\maketitle

\begin{abstract}

Accurate building height estimation is key to the automatic derivation of 3D city models from emerging big geospatial data, including Volunteered Geographical Information (VGI). However, an automatic solution for large-scale building height estimation based on low-cost VGI data is currently missing. The fast development of VGI data platforms, especially OpenStreetMap (OSM) and crowdsourced street-view images (SVI), offers a stimulating opportunity to fill this research gap. In this work, we propose a semi-supervised learning (SSL) method of automatically estimating building height from Mapillary SVI and OSM data to generate low-cost and open-source 3D city modeling in LoD1. The proposed method consists of three parts: first, we propose an SSL schema with the option of setting a different ratio of "pseudo label" during the supervised regression; second, we extract multi-level morphometric features from OSM data (i.e., buildings and streets) for the purposed of inferring building height; last, we design a building floor estimation workflow with a pre-trained facade object detection network to generate "pseudo label" from SVI and assign it to the corresponding OSM building footprint. In a case study, we validate the proposed SSL method in the city of Heidelberg, Germany and evaluate the model performance against the reference data of building heights. Based on three different regression models, namely Random Forest (RF), Support Vector Machine (SVM), and Convolutional Neural Network (CNN), the SSL method leads to a clear performance boosting in estimating building heights with a Mean Absolute Error (MAE) around 2.1 meters, which is competitive to state-of-the-art approaches. The preliminary result is promising and motivates our future work in scaling up the proposed method based on low-cost VGI data, with possibilities in even regions and areas with diverse data quality and availability. Data and code supporting this paper are publicly available in (\url{https://github.com/bobleegogogo/building_height}).
\end{abstract}
\section{Introduction}

For decades, the world has been comprehensively mapped in 2D, however a vertical dimension remains underexplored despite its huge potential, which is even more critical in Global South areas due to inherent mapping inequality and diverse data availability. Mapping human settlements as a 3D representation of reality requires an accurate description of vertical dimension besides the 2D footprints and shapes \cite{kolbe2008citygml, fan2012three, goetz2013towards, biljecki2014height, li2020exploration}. Such 3D representation of human settlements is of significant importance in many aspects, for instance, quiet and shadow routing \cite{ wang2020quiet}, environmental exposure modeling \cite{apte2017high, yuan_knowledge_2022, tost2019neural}, architecture and city planning \cite{resch2016impact,wu2021roofpedia} and population capacity estimation \cite{wurm2011object,kubanek2014capacities}. However, it remains challenging to derive low-cost and open-source 3D representation of buildings at scale. In this paper, with "low-cost", we mainly refer to the cost of data acquisition in 3D building modeling.

Given existing methods of photogrammetry and remote sensing, 3D city reconstruction is still a high-cost and time-consuming task, which mostly requires extensive expert knowledge and a large amount of geospatial data (e.g., cadastral data, airborne photogrammetry data). This fact will certainly increase the difficulty of ordinary stakeholders and city governments with limited funding in establishing 3D city modeling systems for their well-being demands. Fortunately, the increasing availability of Volunteer Geographic Information (VGI) together with crowdsourcing technology \cite{Goodchild2007} has provided a low-cost and scalable solution of mapping our world even in a 3D representation. OpenStreetMap (OSM), as the most successful VGI project, was considered as a valuable global data source for creating large-scale 3D city models \cite{goetz2013towards,fan2016modelling}. For instance, in \cite{esch2020towards}, a joint processing method of OSM and mutli-sensor remote sensing data (e.g., TanDEM-X and Sentinel-2) was developed to generate large-scale 3D urban reconstruction; Milojevic-Dupont et al \cite{milojevic2020learning}. demonstrated the capability of accurate building height prediction purely based on morphometric features (or urban forms) extracted from OSM data (e.g., building and street geometry).

Moreover, several recent works in \cite{zhang2021vgi3d} and \cite{pang20223d} highlight the huge potential of low-cost street-view images (SVI) in increasing the efficiency of large-scale 3D city modeling. The idea is intuitive as SVI provides a low-cost and close-range observation of urban buildings, therefore contains key information needed for 3D reconstruction, such as facade elements, shapes, and building heights. Given the fast development of geospatial machine learning and artificial intelligence (GeoAI) \cite{janowicz2020geoai}, automatic interpretations of SVI have become more efficient than ever before. Hence, the geospatial ML method, which can integrate building height information derived from SVI with existing 2D building footprints from OSM, presents a promising solution for creating large-scale and open-source 3D city models. 

\begin{figure*}[!htbp]
\centering
  \includegraphics[width=\textwidth]{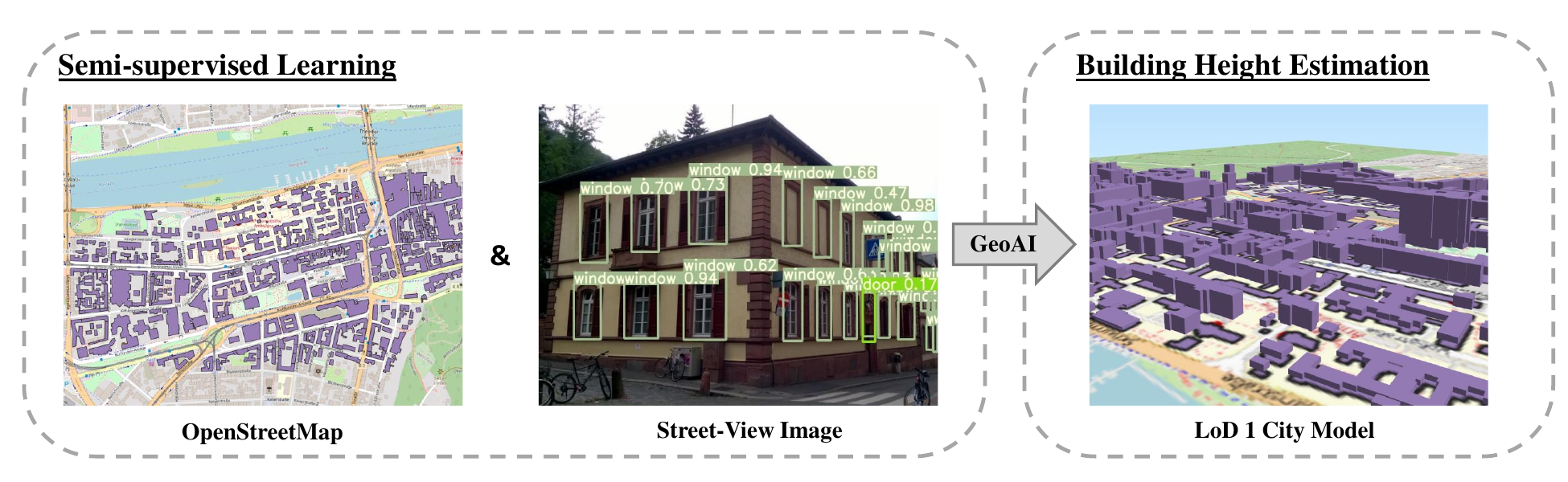}
  \caption{An overview of building height estimation via semi-supervised learning from OpenStreetMap data and street-view images. }
  \label{fig:overview}
\end{figure*}

In this paper, we propose a semi-supervised learning (SSL) method (as shown in Figure \ref{fig:overview}) to accurately estimate building height based on open-source SVI and OSM data. As a case study, we implement the proposed method by training three different machine learning (ML) models, namely Random Forests (RF), Support Vector Machine (SVM), and Convolutional Neural Network (CNN), in the city of Heidelberg, Germany. Specifically, we first extract multi-level urban morphometric features from existing OSM data (i.e., buildings, streets, street blocks) as a feature space to the regression of building height, then we collect SVI with metadata via the Mapillary platform (\url{https://www.mapillary.com}) and design a   building floor estimation workflow with a pre-trained facade object detection network to generate "pseudo label" for the SSL of building height estimation models. As a result, we create an open-source LoD1 3D city models for selected areas in Heidelberg using the low-cost SVI data and OSM 2D building footprints. 

\section{Related Work}

\subsection{Building Height Estimation}
Existing methods of building height estimation generally rely on Light Detection and Ranging (LiDAR) \cite{gongICESatGLASData2011,parkCreating3DCity2019}, Synthetic Aperture Radar (SAR) \cite{liDevelopingMethodEstimate2020}, and high-resolution remote sensing image data \cite{liuIM2ELEVATIONBuildingHeight2020}. In these data sources, LiDAR data provides highly accurate information of building height but is difficult to estimate building height in large scale, considering its collection cost.  For SAR, the estimation result is often affected by the mixture of different microwave scattering, thus have high uncertainties \cite{sunLargescaleBuildingHeight2019}. To avoid these problems, many researchers also investigate remote sensing image data. For these methods, considering that remote sensing image data does not contain 3D information directly, existing works select stereo/multi-view images as the data source to achieve the estimation of building height \cite{alobeid2009building, caoDeepLearningMethod2021,zhangBuildingHeightExtraction2022}. 

However, although SAR and remote sensing image data have a relatively low collection cost than LiDAR data, the complex data processing of these data source causes their high time and labor costs. Compared with these three data, SVI data and 2D building footprint data are easier and cheaper to be collected and processed, especially with the support of VGI (e.g., Mapillary and OpenStreetMap). There have been some early efforts to estimate building height based on these new data sources. Biljecki et al. \cite{biljecki2017generating}, Milojevic-Dupont et al. \cite{milojevic2020learning}, and Bernard et al. \cite{bernardEstimationMissingBuilding2022} proposed several methods based on RF or other ML approaches to analyze the relationship between building heights and their features (such as building area and type), and finally achieve the building height estimation from 2D footprint data. Yan and Huang \cite{yanEstimationBuildingHeight2022} proposed a deep learning-based method to estimate building height from SVI. Zhao et al. \cite{zhaoCBHECornerbasedBuilding2019} combined 2D building footprints and SVI to estimate building heights, which also used deep learning technology. These methods also achieved good performance but require a large amount of training data, which limits their generalization and practicality. Currently, there is little work on how to accurately estimate building height from 2D building footprint and SVI with only limited training data.

\subsection{VGI and 3D Building Models}
CityGML is a well-known international standard for 3D building modeling. In CityGML 2.0, 3D building models are divided into five levels of detail (LoDs). In LoD0, only the 2D footprint information is involved in the model. In LoD1, the LoD0 model is extruded by their building heights, and the obtained cuboid after extrusion are the LoD1 model. In LoD2, the 3D roof structure information is added into the LoD2 model. The LoD3 model further contains the facade element information, such as windows and doors. The LoD4 model is more complicated and contains both external and internal building elements. To meet the requirements of the abovementioned CityGML standard, many cities like New York, Singapore, and Berlin have created and freely released 3D city models with different LoDs in the past years. However, most of these 3D city building models are constructed in LoD1 or LoD2 for urban area, while large-scale and fine-grained (LoD3 and LoD4) models with semantic information are hardly available for cities with limited funding in establishing their own 3D city modelling systems. Hence, that is the main motivation of this work to provide a low-cost and open-source solution of creating large-scale 3D city models (e.g., first in LoD1).

Early work in \cite{goetz2013towards} highlighted that OSM, as a crowdsourced VGI data source, can be combined with international standards of the Open
Geospatial Consortium (OGC) to effectively create CityGML models in LoD1 and LoD2. Recently, Zhang et. al \cite{zhang2021vgi3d} proposed a web-based interactive system, namely VGI3D, as a collaborative platform to collect 3D building models with fine-grained semantic information in a crowdsourcing approach. In this work, we aim to further investigate the potential of low-cost VGI data sources, especially OSM data and crowdsourced SVI, in generating LoD1 3D city models via automatic building height estimation with only limited training data.

\section{Methodology}

\begin{figure*}[!htbp]
\centering
  \includegraphics[width=\textwidth]{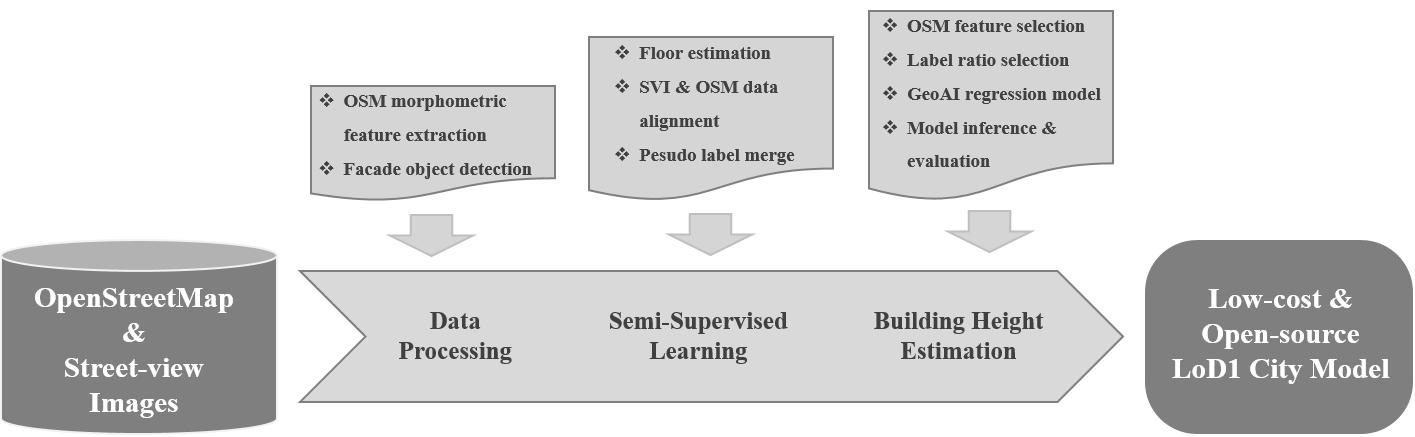}
  \caption{The methodological workflow of automatic building height estimation from OpenStreetMap data and street-view images. }
  \label{fig:methodology}
\end{figure*}

The proposed method of automatic building height estimation mainly consists of three parts: (1) an SSL schema for height regression, (2) OSM morphometric feature extraction, and (3) building floor estimation based on the SVI. Figure \ref{fig:methodology} shows the methodological workflow of automatically generating open-source 3D city modeling (i.e., LoD1 city model) via the proposed SSL method. In the rest of this section, we will elaborate on the details of this design.

\subsection{Semi-supervised Learning Schema}

In traditional supervised learning, one relies on labelled data to build the prediction model. However, such a labelling process is mostly time consuming, labour demanding, and difficult to scale up. Therefore, the capability of learning from unlabeled data is a desirable feature to overcome this challenge. In this context, Semi-supervised learning (SSL) is a promising technique to accommodate the lack of labeled data by allowing the model to integrate part of unlabeled data during the supervised model training \cite{zhu2005semi, jing2020self}. To be noticed, the SSL herein is different from self-supervised learning, which does not rely on any ground truth labels during the training process. A common way of implementing SSL is to generate "pseudo label" from the data itself or even auxiliary data \cite{lee2013pseudo}, which can be then merged with existing labelled data to boost model performance. Following this concept, we design an SSL schema with the option of defining different ratio of "pseudo label" during the supervised regression of building height.

The proposed SSL schema is tasked with estimating building heights ($h$) based on a list of morphometric features $x = \langle x_1, \dots, x_m \rangle$ extracted from diverse scales of OSM data (e.g., individual building footprint, street network, street block, etc.), where $m$ refers to the total number of features. In this context, the task of building height estimation can be formulated as a multifactor regression task in the following mathematic form:

\begin{equation}	
	h_{\Theta}(x) = \sum_{i=0}^{m} \Theta_i x_i
	\label{eq:lr}
\end{equation}
where $\Theta = \langle \Theta_1, \dots, \Theta_m \rangle$ is the corresponding regression coefficients. More importantly, the regression target value of building heights $h$ comes from the following two parts:

\begin{equation}	
	h =  (1-a)*h_{Raw} + a*h_{SSL}
\end{equation}
Where $a$ is the ratio of "pseudo label" ($h_{SSL}$) obtained from automatic facade parsing of Mapillary SVI. We will elaborate on this later in Section \ref{SVI}, while it is sufficient to understand that besides available training label (i.e., known building heights) the model can also benefit from SSL labels which are extracted from large-scale and open-source SVI in an automatic and unsupervised method.

To build the model for accurate building height estimation, we train a classic supervised regression model of finding the optimal regression coefficients with gradient descent and optimizing a loss function of Mean Square Error (MSE) in the following format:  

\begin{equation}	
		\mathcal{L}^{MAE}_{\Theta^{*}} = \arg \min_{\Theta} \frac{1}{N} \sum_{i=1}^{N} \parallel \hat{h}_i - h_i \parallel
	\label{eq:loss}
\end{equation}
where $\mathcal{L}_{MAE}$ and $\Theta^{*}$ refer to the loss function and the optimal coefficients set, respectively, and $N$ is the number of training samples ($h_{SSL}$ and $h_{Raw}$).

The design of SSL is concise and model-independent, which means in case we can keep feeding the ML models with "pseudo label" ($h_{SSL}$) of building height extracted from SVI, the regression task can be tackled with diverse ML models. In this paper, we demonstrate the capability of these three ML models (i.e., RF, SVM, and CNN) in estimating building height in a typical western European city, so to say the city of Heidelberg, Germany.

\subsection{OSM Morphometric Feature Extraction}

Intensive existing works have confirmed the excellent capability of multi-level morphological features (or urban-form features) in predicting key attributes (e.g., height, function, energy consumption, etc.) of buildings and streets from an urban analytic perspective \cite{milojevic2020learning, biljecki2022global}. 

To infer building height, we implement a range of morphometric features extracted from OSM at three different levels, namely building-level, street-level, and street block-level, as shown in Table \ref{tab:OSM_features}. In total, we calculate 129 morphometric features based on OSM data (i.e., individual building footprints and street networks) to construct their spatial and geometric relationships (e.g., spatial vicinity and compactness of street-blocks). More specifically, we elaborate on the details of OSM morphometric features (in three distinct levels) as follows:

\textbf{Building-level}: Considering the hidden information from the building footprint itself, we calculate 9 features such as footprint area, perimeter, circular compactness, convexity, orientation and length of wall shared with other buildings. The intuition herein is that such building-level features can provide explicit and implicit information about the footprint shape (e.g., compactness and complexity), which contributes to estimating building heights. For instance, it was reported that a higher building generally consists of a large net internal area, and vice versa \cite{biljecki2017generating}. In addition, since buildings are mapped differently in OSM (e.g., one building in several polygons or several buildings in one polygon), we simplify this data quality issue by considering each polygon as a single building, while future work is definitely needed in investigating the impact of how individual buildings are presented in OSM.

\textbf{Street-level}: Besides morphometric features of the building footprint itself, the street network surrounding a building can be informative in estimating building height. For instance, a high density (or compactness) of streets can imply more high-story buildings in order to accommodate a potentially higher number of residents. Therefore, we calculate 9 features based on the spatial relationship of buildings and their closest streets and road intersections, such as length, average width, distance to the building, local closeness, betweenness and centrality, etc.  

\textbf{Street block-level}: Furthermore, we generate morphological tessellations based on the OSM street network. This tessellation representation and its interaction with roads and buildings were included in the design of the feature space (8 features). The motivation is straightforward, as a preliminary assumption is that buildings in the same block are more likely to be of a similar height.

Moreover, to capture the spatial auto-correlation in the OSM data, we extend these three levels of OSM morphometric features by considering their second-order features (e.g., total, average, and standard deviation) in the neighbourhood (i.e., within 20, 50, and 500 meters buffers). As for the implementation, we rely on the open-source Python software toolkit called momepy v.0.5.1 to calculate these features. For a complete list of OSM morphometric features, please refer to the GitHub repository (\url{https://github.com/bobleegogogo/building_height}).

\renewcommand\arraystretch{1.1}
\begin{table*}[!htbp]
\centering
\centering
\caption{List of OSM morphometric features extracted at building-level, street-level, and street-block level. }
\label{tab:OSM_features}
\begin{adjustbox}{width=\textwidth}
\begin{tabular}{c l l c c c c c c c c c}

\multicolumn{1}{c}{Level}
&\multicolumn{1}{c}{Group}
&\multicolumn{1}{c}{Features}
&\multicolumn{1}{c}{Count} \\
\hline

\multirow{4}{*}{\textbf{Building}}
& building footprint & e.g., area, perimeter, conversity etc.& \textbf{9} \\
& buildings within 50m  & e.g., total/average/standard deviation of area,perimeter,conversity etc.& \textbf{18} \\
& buildings within 200m  & e.g., total/average/standard deviation of perimeter, etc.& \textbf{18} \\
& buildings within 500m  & e.g., total/average/standard deviation of conversity, etc.& \textbf{18} \\
\hline

\multirow{4}{*}{\textbf{Street}}
& closest streets and intersections & e.g., length, closeness and distances to the intersection & \textbf{9} \\
& closest streets and intersections within 50m  & e.g., total/average/standard deviation of distances to the closest intersection & \textbf{11} \\
& closest streets and intersections within 200m  & e.g., total/average/standard deviation of distances to the closest intersection & \textbf{11} \\
& closest streets and intersections within 500m  & e.g., total/average/standard deviation of distances to the closest intersection & \textbf{11} \\
\hline

\multirow{3}{*}{\textbf{Street-block}}
& street-block itself & e.g., area, convexity, orientation, corner count, etc.& \textbf{8} \\
& buildings in blocks  & e.g., count, total/average/standard deviation of the area of building & \textbf{4} \\
& street-block within 50, 200 and 500m  & e.g., total/average/standard corner count, area of blocks etc.& \textbf{14} \\
\hline
\multirow{1}{*}{\textbf{Total features}}
&  & & \textbf{129} \\

\end{tabular}
\end{adjustbox}
\end{table*}

\subsection{Building Floor Estimation from Street-Level Images} \label{SVI}

Inspired by the work of automatic facade parsing in \cite{kong2020enhanced}, we develop a building floor estimation workflow based on automatic facade parsing and urban architecture rules. In short, we aim to generate the estimation of building floor or height (by multiplying an average floor height) as the "pseudo label" to guide ML regression models with the aforementioned OSM morphometric features as covariates. Figure \ref{fig:SVI} illustrates the developed method of building floor estimation based on SVI. To explain the developed method in more detail, we elaborate on three main steps as follows:

\begin{figure*}[!htbp]
\centering
  \includegraphics[width=\textwidth]{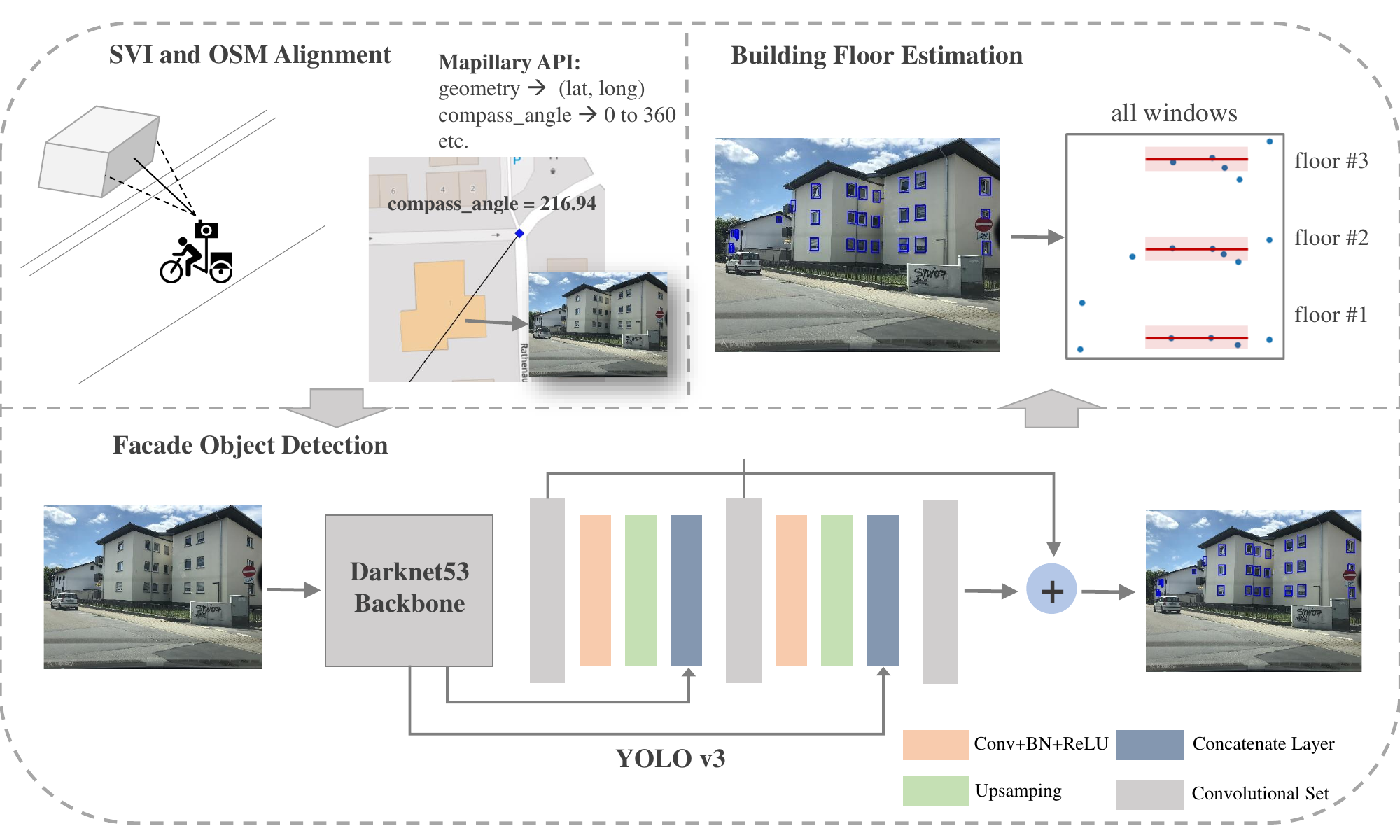}
  \caption{Three steps of building floor estimation from street-level images: (1) aligning SVI and OSM building; (2) facade parsing using object detection; (3) generating "pseudo label" by building floor estimation.}
  \label{fig:SVI}
\end{figure*}

\textbf{SVI and OSM building alignment}: As the first step, we download existing SVI from Mapillary via their open-source image API, where each SVI record consists of geotagged coordinates of the camera during a trip sequence and additional metadata information (Table \ref{tab:api_endpoints}), especially the compass angle of the camera direction (i.e., 0 to 360 degrees). This compass angle together with geotagged coordinates of the camera is key for aligning SVI with an individual OSM building. To this end, we apply a simple ray-tracing method to determine their relationship and assign the selected Mapillary SVI to its corresponding OSM building footprints (see Figure \ref{fig:SVI}). Currently, we manually select Mapillary images which cover the complete facade of a building without being blocked by vegetation and cars, while future work is needed to automate this selecting process. A possible solution is to apply semantic segmentation approaches and ensure the skyline and ground are both visible within a single SVI.

\renewcommand\arraystretch{1.1}
\begin{table}[!htbp]
\centering
\centering
\caption{Selected metadata of SVI from the Mapillary Image API Endpoints}
\label{tab:api_endpoints}
\begin{adjustbox}{width=\textwidth}
\begin{tabular}{l l l l l l c c}
\hline

\multicolumn{1}{c}{Fields}
&\multicolumn{1}{c}{Data Format}
&\multicolumn{1}{c}{Description}\\
\hline

\multirow{1}{*}{computed\_geometry}
& GeoJSON Point & latitude and longitude after running image processing.  \\
\multirow{1}{*}{computed\_compass\_angle}
& float & compass angle of the camera direction. \\
\multirow{1}{*}{computed\_altitude}
& float & altitude after running image processing, from sea level. \\
\multirow{1}{*}{computed\_rotation}
& enum & corrected orientation of the image, refer to OpenSfM definitation. \\
\multirow{1}{*}{camera\_type }
& enum & type of camera projection: "perspective", "fisheye", "equirectangular". \\
\multirow{1}{*}{captured\_at}
&  timestamp & capture time of the camera. \\
\multirow{1}{*}{camera\_parameters}
&  array of float & focal length, k1, k2 of the camera. \\
\multirow{1}{*}{exif\_orientation}
& enum & orientation of the camera as given by the Exif tag. \\

\hline
\multicolumn{3}{l}{Note: All fields refer to Mapillary API Version 4.}

\end{tabular}
\end{adjustbox}
\end{table}

\textbf{Facade object detection}: There are two common approaches in measuring building heights: either estimating absolute metrics (e.g., meters) or counting the floor number. As for accurately inferring the floor number, key features (e.g., window, balcony, and door) and their layout in the building facade play a key role \cite{biljecki2017generating}. Herein, we aim to detect these key features from street-level Mapillary imagery via the facade parsing technique. To this end, we follow the deep learning method developed in \cite{kong2020enhanced} for automatic facade parsing from the SVI data. Specifically, we use a pre-trained one-stage object detection network, namely YOLO v3 \cite{redmon2018yolov3} (with the Darknet53 backbone), for the purpose of fast and accurate facade object detection. Herein, the facade object detection has been pre-trained on a facade semantic dataset called FaçadeWHU \cite{kong2020enhanced}, thus could be directly applied to detect key facade features (e.g., window, balcony, and door) from the Mapillary SVI collected in Heidelberg without further training. As a result, the detected facade features are saved as a list of objects and their image coordinates. 

\textbf{Building floor estimation}: Based on facade object detection results, we then apply a rule-based approach to determine the floor number in order to estimate the height of corresponding OSM buildings. Specifically, we first group facade objects (i.e., windows and doors) with their vertical coordinates and calculate the difference between each two neighbored elements, next k-mean clustering (with k=2) is used to find the clusters where objects are aligned vertically with each other, which results in a floor number estimation by counting the number of windows. By considering an average floor-to-floor height (i.e., 2.5 meters for residential or 3.5 meters for commercial), we can then derive the building height information from the SVI data, and use it as an SSL training label ($h_{SSL}$) to train the ML regression model on OSM morphometric features.

\section{Preliminary Result} \label{Experiment}

\subsection{Case Study}
As a case study, we implemented and tested the proposed method (Figure \ref{fig:methodology}) in a classic western European city, namely the city of Heidelberg, Germany by considering Heidelberg was relatively well-mapped in OSM. Moreover, the reference data ($h_{Raw}$) of building heights obtained from the City of Heidelberg is also available, where building eaves heights (as we aim at LoD1 model for now) were recorded and spatially joined with OSM building footprints.

We extracted the latest OSM data (buildings and streets) via the ohsome API, which is built on the OpenStreetMap History Database (OSHDB) \cite{raifer2019oshdb}. Herein, the ohsome API enables us to trace back to even historical OSM data, which can potentially contribute to more intrinsic features (e.g., the curve of nodes or contributions density). However, this goes beyond the scope of this paper. In this work, we calculated 129 morphometric features for 16,089 building footprints within the city of Heidelberg, which were used to train three types of ML regression models, specifically RF with 1000 trees, SVM with RBF kernel, and a three-layer dense CNN, to estimate building heights.

Regarding the SVI data, we followed the method described in Figure \ref{fig:SVI} by manually choosing 308 street-level Mapillary images and aligning them with 308 corresponding OSM building footprints by considering the SVI metadata. Then, we estimated their floor number and further converted them into building heights by multiplying an average height of 2.5 meters for residential buildings and 3.5 meters for commercial and public buildings \cite{chun2012two}. Herein, we manually verified the building function for these 308 SVI and their corresponding OSM building footprints. Although it is possible to automate this process with OSM data \cite{fan2014estimation, atwal2022predicting}, the prediction of building functions is beyond the scope of this paper. Despite its limitation, the proposed method provides a promising and low-cost solution to create open-source 3D city models (LoD1) by consuming only VGI data sourced (i.e., OSM and SVI) with a flexible SSL schema. 

\subsection{Experimental Result}

In our case study, we conduct two comparative analysis to evaluate the capability of our SSL method w.r.t mainly two variables: first, the different OSM morphometric features, second, the different ratio of "pseudo label" during SSL training, by comparing the regress performance among three ML regression models (e.g., RF, SVM, and CNN). 

\textbf{Height estimation with different OSM features}: To validate multi-level morphometric features extracted from OSM, Table \ref{tab:regression_result} compares the regression performance of three ML models (RF, SVM, and CNN) using two different levels of morphometric features (i.e., 64 building-level features and all 129 features). Herein, we set a split ratio (between training and testing samples) of 0.7 on the reference data and calculate three common regression metrics (MAE, RMSE, and $R^2$, all in meters) for the evaluation purpose. An important finding is that the integration of street and street-block features leads to an incremental boosting in the model performance, though this is less significant in the case of CNN. Though in the case of SVM, more features seem to be not helpful. A potential reason can be attributed to a potential effect of the curse of dimensionality. In short, an average MAE of around 2.3 meters (RF with 129 features), which is less than the average height of a single floor, confirms the feasibility of accurately estimating building height only from OSM morphometric features. This result encourages us to incorporate these OSM morphometric features with the proposed SSL method to better create large-scale and open-source 3D city models.
\renewcommand\arraystretch{1.1}
\begin{table}[!htbp]
\centering
\centering
\caption{Preliminary results of estimating building heights with different OSM features and regression models.}
\label{tab:regression_result}
\begin{adjustbox}{width=0.8\textwidth}
\begin{tabular}{c c c c c c c}

\hline
\multicolumn{1}{c}{}
&\multicolumn{2}{c}{RF}
&\multicolumn{2}{c}{SVM}
&\multicolumn{2}{c}{CNN} \\
\hline
Feature & 64 & 129 & 64 & 129 & 64 & 129 \\

\hline
\multirow{1}{*}{MAE}
&  2.58  & \textbf{2.38} & 2.89 &  2.91 & 2.78 & 2.67 \\

\multirow{1}{*}{RMSE}
&  3.55  & \textbf{3.34} & 3.89 &  3.91 & 3.71 & 3.62\\

\multirow{1}{*}{$R^2$}
&  0.2235  & \textbf{0.3140} & 0.0681 &  0.0567 & 0.1515 & 0.1929 \\
\hline
\end{tabular}
\end{adjustbox}
\end{table}

\textbf{SSL with different ratio of "pseudo label"}: Based on the workflow described in Figure \ref{fig:SVI}, we are able to collect 308 SVI from Mapillary and extract "pseudo label" via facade object detection, then associate these height values with their corresponding OSM building footprints. To test the impact of different SSL ratio, we set up three training sets: 1) to use only estimated heights from SVI (SVI) as an aggressive scenario of SSL; 2) to randomly select 308 OSM buildings and retrieve their heights from the reference data to simulate the fully supervised scenario (RAW); 3) to merge the "pseud label" with reference heights thus have a balance SSL training set (i.e., 308 each for SVI and RAW). In addition, a valuation set with 2,000 buildings randomly extracted from the reference data is considered given the limited number of training labels. Table \ref{tab:regression_result_ratio} shows the numerical results using different ratio of "pseudo label" (e.g., SVI, RAW, and SSL) and three ML regression models (with 129 features). Although the "pseudo label" (SVI) still leads to the largest error (w.r.t MAE and RMSE) in all three regression models, the "pseudo" height extracted from SVI is indeed informative for building height regression, more importantly, it is beneficial when merging with existing labels. Therefore, the quantitative result listed in Table \ref{tab:regression_result_ratio} confirms that the proposed SSL method is effective and efficient in extracting "pseudo" training information from crowdsourced SVI data, which largely boosts the estimation accuracy using all three different ML regression models. In future work, it would be interesting to further investigate how different building types (e.g., residential or commercial, one-floor or multi-floor) can affect the accuracy of building height estimation. 

\renewcommand\arraystretch{1.1}
\begin{table}[!htbp]
\centering
\centering
\caption{Preliminary results of  estimating building heights with different training sets and regression models.}
\label{tab:regression_result_ratio}
\begin{adjustbox}{width=\textwidth}
\begin{tabular}{c c c c c c c c c c c c c c c c}

\hline
\multicolumn{1}{c}{}
&\multicolumn{3}{c}{RF}
&\multicolumn{3}{c}{SVM}
&\multicolumn{3}{c}{CNN}\\
\hline
Label & SVI & RAW & SSL & SVI & RAW & SSL & SVI & RAW & SSL \\
\hline
\multirow{1}{*}{MAE}
& 2.75 & 2.67 & \textbf{2.07} & 2.93 & 2.89 & 2.20 & 3.23 & 3.03 & 2.72 \\

\multirow{1}{*}{RMSE}
& 3.85 & 3.80 & \textbf{2.99} & 3.99 & 3.87 & 3.47 & 4.11 & 3.99 & 3.71 \\

\multirow{1}{*}{$R^2$}
& 0.2302 &  0.2210 & \textbf{0.5368} &  0.1726  & 0.0315 & 0.3735 &  0.0241 & 0.1718 & 0.2458 \\
\hline

\end{tabular}
\end{adjustbox}
\end{table}

\begin{figure}[!htbp]
\centering
  \includegraphics[width=0.8\textwidth]{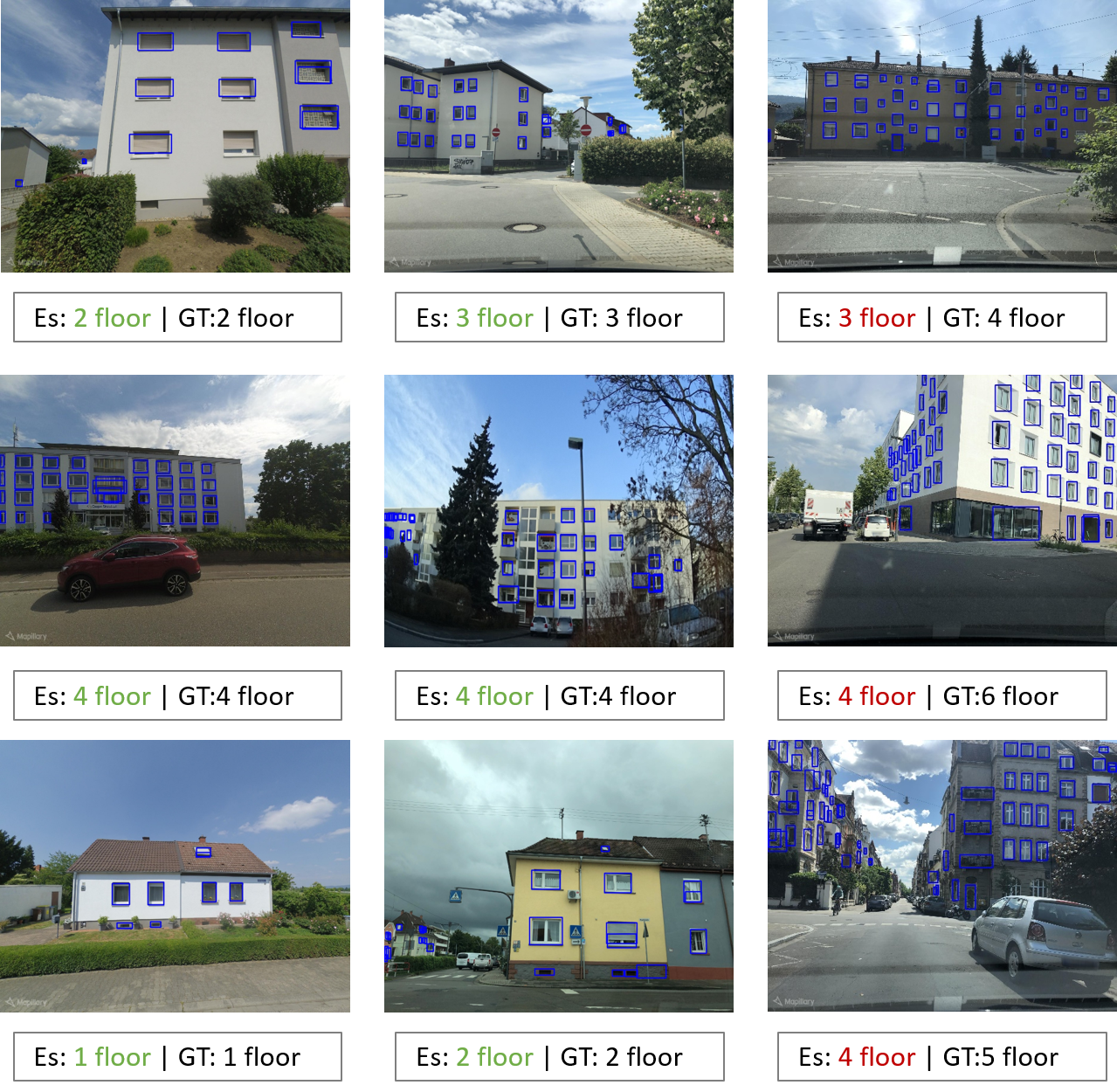}
  \caption{Selected examples of facade object detection and floor number estimating from Mapillary images in Heidelberg.}
  \label{fig:facade_parsing}
\end{figure}

Regarding the generation of "pseudo label", Figure \ref{fig:facade_parsing} shows selected examples of building floor estimations from Mapillary SVI in Heidelberg. One can observe that for lower floor numbers in case the captured facade is complete, the model works in a sensible way. However, we encountered several challenging cases when the building facade is not complete or the layout of windows (e.g., dormer windows) is difficult to be grouped by our floor estimation rules. In this context, future work is needed to develop a more robust method of extracting and distinguishing related features from SVI, such as roof types, dormer windows, and building functions, which can be helpful to generate more reliable "pseudo labels" for the SSL method.

\section{Discussion} \label{Future}

\begin{figure*}[!htbp]
\centering
  \includegraphics[width=0.9\textwidth]{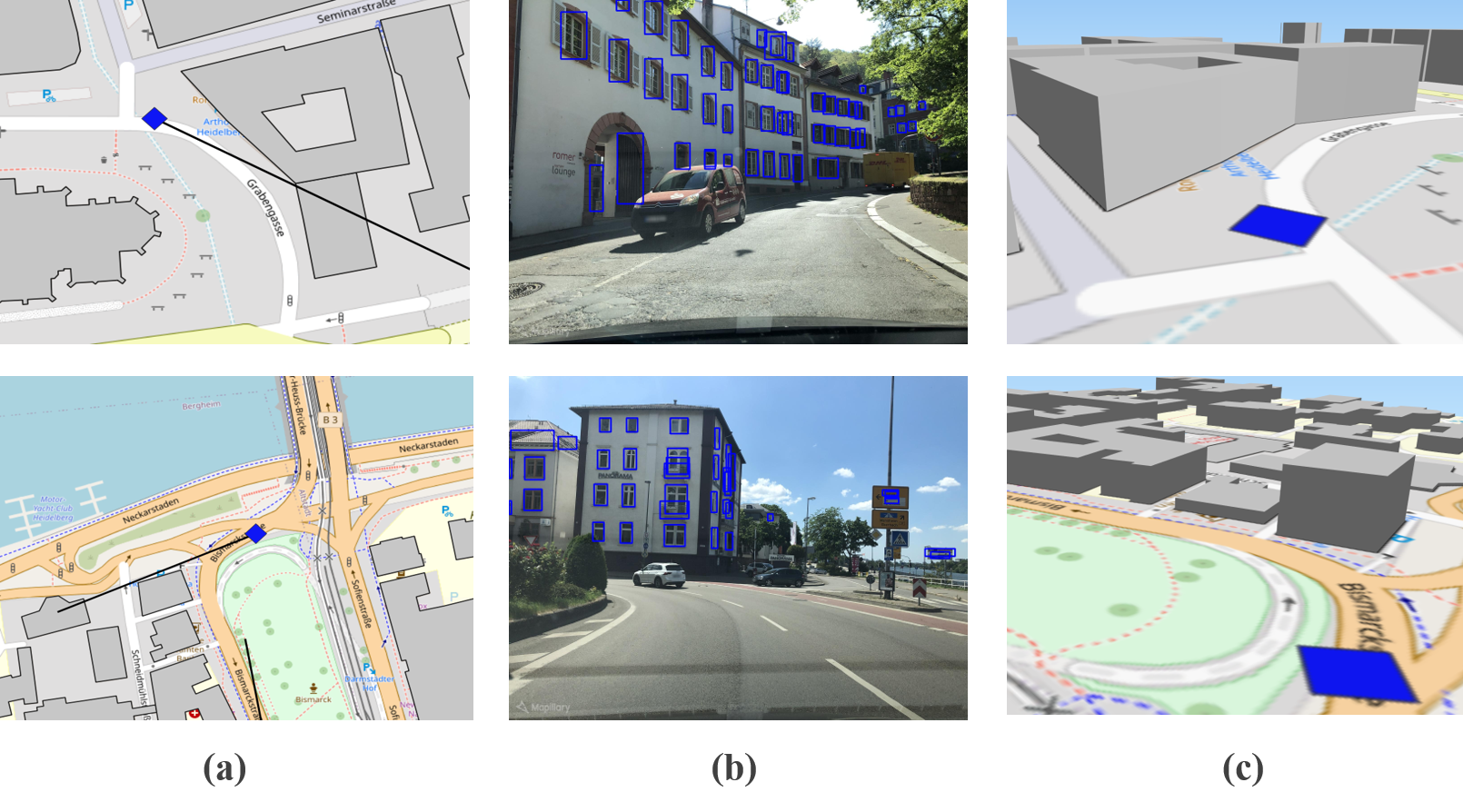}
  \caption{The creation of a LoD1 3D city model using SVI and OSM data in the old town of Heidelberg. (a) OSM data with SVI metadata; (b) SVI with face object detection results; (c) LoD1 model with estimated building heights.}
  \label{fig:SVI_3D}
\end{figure*}

In Figure \ref{fig:SVI_3D}, we demonstrate a 3D city model in LoD1 for selected buildings in the old town of Heidelberg, which is created using the proposed SSL method based on SVI (Figure \ref{fig:SVI_3D} (b)) and OSM building footprints (Figure \ref{fig:SVI_3D} (a)). In future work, we aim to refine this method by addressing the aforementioned limitations and comparing the estimated one with official LoD1 city models in selected cities. 

Our preliminary result echoes the findings in \cite{kong2020enhanced} and \cite{milojevic2020learning} to a certain extent. More importantly, the SSL method will make our method in principle even more flexible and easy-to-apply in areas where the availability of training data (e.g., existing building heights) is limited or difficult-to-access. For instance, in most developed countries, 3D city models can be established using e.g., Digital Terrain Model (DTM), however the acquisition of large-scale and accurate DTM data remains costly and time-consuming. In this context, the proposed method provides a solution to directly harness existing crowdsourced VGI data (OSM and SVI) for 3D city modeling without additional data acquisition (e.g., DTM). Therefore, the "low-cost" herein mainly refers to the cost of traditional data acquisition methods w.r.t building height information. Despite the high potential, we identify several limitations to be addressed in future work: 

\begin{itemize}
    \item It is key to improve the building floor estimation workflow in terms of accuracy and speed, with which more "pseudo labels" can be extracted and used for SSL. For instance, the current SVI selection is done manually to ensure complete coverage of a building facade without being blocked by vegetation and cars, while this process can be automated using a semantic segmentation approach to improve the efficiency of generating high-quality "pseudo labels" at scale. Moreover, OSM data itself may contain information about building height (\textit{"building:levels=* or height=*"}) as well, which could be a helpful source to get more training data into the SSL method. 
    \item Despite its low-cost and open-source nature, the quality aspect of VGI data (i.e., OSM and SVI data) remains under-quantified in this work, but certainly deserves a careful and decent treatment when applied to different countries or cities in the world \cite{li2022improving}. For instance, the positional error and obstruction in SVI can significantly hinder the existing floor estimation approach. In addition, one needs to investigate how many SVI images are needed to have a reasonable spatial coverage of a study area to ensure the effectiveness of the SSL method.
    \item The ML regression models used in this work are based on a 1D vector feature space (up to 129 different features). However, a more sophisticated method is needed to encode the spatial relationship among buildings. For example, one option is to apply a graph CNN \cite{yan2019graph} as a spatial-explicit building height regressor.
    \item It is still unclear how different architecture types (e.g., roof type, construction age, building function) and city styles (e.g., low-rise, medium-rise, or high-rise) will affect the effectiveness and accuracy of our SSL method.
\end{itemize}

\section{Conclusion} \label{Conclusion}

In this paper, we present a semi-supervised learning (SSL) method of automatic building height estimation by integrating crowdsourced street-level images (SVI) with multi-level morphometric features extracted from the OpenStreetMap (OSM) data. In this context, we design a workflow to convert facade object detection results from Mapillary SVI into "pseudo label" of building heights for three different ML regression models. As a case study, we validate the proposed SSL method in the city of Heidelberg, Germany, and the preliminary result looks very promising. However, the varying quality of volunteered geographical information (VGI) data, cultural and city-wise differences in the morphological features used, and the varying availability of SVI, all lead to certain limitations of such an SSL method. Our future work will focus on tackling these limitations and provide a robust and scalable solution of large-scale and open-source 3D city modeling purely based on low-cost VGI data.


\bibliographystyle{plainurl}
\bibliography{lipics-v2019-sample-article}

\begin{thebibliography}{10}

\bibitem{alobeid2009building}
A~Alobeid, K~Jacobsen, and C~Heipke.
\newblock Building height estimation in urban areas from very high resolution
  satellite stereo images.
\newblock In {\em ISPRS Hannover Workshop}, volume~5, pages 2--5, 2009.

\bibitem{apte2017high}
Joshua~S Apte, Kyle~P Messier, Shahzad Gani, Michael Brauer, Thomas~W
  Kirchstetter, Melissa~M Lunden, Julian~D Marshall, Christopher~J Portier,
  Roel~CH Vermeulen, and Steven~P Hamburg.
\newblock High-resolution air pollution mapping with google street view cars:
  exploiting big data.
\newblock {\em Environmental science \& technology}, 51(12):6999--7008, 2017.

\bibitem{atwal2022predicting}
Kuldip~Singh Atwal, Taylor Anderson, Dieter Pfoser, and Andreas Z{\"u}fle.
\newblock Predicting building types using openstreetmap.
\newblock {\em Scientific Reports}, 12(1):19976, 2022.

\bibitem{bernardEstimationMissingBuilding2022}
J{\'e}r{\'e}my Bernard, Erwan Bocher, Elisabeth Le~Saux~Wiederhold, Fran{\c
  c}ois Leconte, and Val{\'e}ry Masson.
\newblock Estimation of missing building height in {{OpenStreetMap}} data: A
  {{French}} case study using {{GeoClimate}} 0.0.1.
\newblock {\em Geoscientific Model Development}, 15(19):7505--7532, October
  2022.
\newblock \href {https://doi.org/10.5194/gmd-15-7505-2022}
  {\path{doi:10.5194/gmd-15-7505-2022}}.

\bibitem{biljecki2022global}
Filip Biljecki and Yoong~Shin Chow.
\newblock Global building morphology indicators.
\newblock {\em Computers, Environment and Urban Systems}, 95:101809, 2022.

\bibitem{biljecki2017generating}
Filip Biljecki, Hugo Ledoux, and Jantien Stoter.
\newblock Generating 3d city models without elevation data.
\newblock {\em Computers, Environment and Urban Systems}, 64:1--18, 2017.

\bibitem{biljecki2014height}
Filip Biljecki, Hugo Ledoux, and JE~Stoter.
\newblock Height references of citygml lod1 buildings and their influence on
  applications.
\newblock In {\em Proceedings. 9th ISPRS 3DGeoInfo Conference 2014, 11-13
  November 2014, Dubai, UAE,(authors version)}. Citeseer, 2014.

\bibitem{caoDeepLearningMethod2021}
Yinxia Cao and Xin Huang.
\newblock A deep learning method for building height estimation using
  high-resolution multi-view imagery over urban areas: {{A}} case study of 42
  {{Chinese}} cities.
\newblock {\em Remote Sensing of Environment}, 264:112590, October 2021.
\newblock \href {https://doi.org/10.1016/j.rse.2021.112590}
  {\path{doi:10.1016/j.rse.2021.112590}}.

\bibitem{chun2012two}
Bumseok Chun and Jean-Michel Guldmann.
\newblock Two- and three-dimensional urban core determinants of the urban heat
  island: A statistical approach.
\newblock {\em Journal of Environmental Science and Engineering B},
  1(3):363--378, 2012.

\bibitem{esch2020towards}
Thomas Esch, Julian Zeidler, Daniela Palacios-Lopez, Mattia Marconcini, Achim
  Roth, Milena M{\"o}nks, Benjamin Leutner, Elisabeth Brzoska, Annekatrin
  Metz-Marconcini, Felix Bachofer, et~al.
\newblock Towards a large-scale 3d modeling of the built environment—joint
  analysis of tandem-x, sentinel-2 and open street map data.
\newblock {\em Remote Sensing}, 12(15):2391, 2020.

\bibitem{fan2012three}
Hongchao Fan and Liqiu Meng.
\newblock A three-step approach of simplifying 3d buildings modeled by citygml.
\newblock {\em International Journal of Geographical Information Science},
  26(6):1091--1107, 2012.

\bibitem{fan2016modelling}
Hongchao Fan and Alexander Zipf.
\newblock Modelling the world in 3d from vgi/crowdsourced data.
\newblock {\em European handbook of crowdsourced geographic information}, 435,
  2016.

\bibitem{fan2014estimation}
Hongchao Fan, Alexander Zipf, and Qing Fu.
\newblock Estimation of building types on openstreetmap based on urban
  morphology analysis.
\newblock {\em Connecting a digital Europe through location and place}, pages
  19--35, 2014.

\bibitem{goetz2013towards}
Marcus Goetz.
\newblock Towards generating highly detailed 3d citygml models from
  openstreetmap.
\newblock {\em International Journal of Geographical Information Science},
  27(5):845--865, 2013.

\bibitem{gongICESatGLASData2011}
Peng Gong, Zhan Li, Huabing Huang, Guoqing Sun, and Lei Wang.
\newblock {{ICESat GLAS Data}} for {{Urban Environment Monitoring}}.
\newblock {\em IEEE Transactions on Geoscience and Remote Sensing},
  49(3):1158--1172, March 2011.
\newblock \href {https://doi.org/10.1109/TGRS.2010.2070514}
  {\path{doi:10.1109/TGRS.2010.2070514}}.

\bibitem{Goodchild2007}
Michael~F. Goodchild.
\newblock Citizens as sensors: The world of volunteered geography.
\newblock {\em GeoJournal}, 69:211--221, 08 2007.
\newblock \href {https://doi.org/10.1007/s10708-007-9111-y}
  {\path{doi:10.1007/s10708-007-9111-y}}.

\bibitem{janowicz2020geoai}
Krzysztof Janowicz, Song Gao, Grant McKenzie, Yingjie Hu, and Budhendra
  Bhaduri.
\newblock Geoai: spatially explicit artificial intelligence techniques for
  geographic knowledge discovery and beyond.
\newblock {\em International Journal of Geographical Information Science},
  34(4):625--636, 2020.

\bibitem{jing2020self}
Longlong Jing and Yingli Tian.
\newblock Self-supervised visual feature learning with deep neural networks: A
  survey.
\newblock {\em IEEE transactions on pattern analysis and machine intelligence},
  43(11):4037--4058, 2020.

\bibitem{kolbe2008citygml}
Thomas~H Kolbe, Gerhard Gr{\"o}ger, and Lutz Pl{\"u}mer.
\newblock Citygml--3d city models and their potential for emergency response.
\newblock In {\em Geospatial information technology for emergency response},
  pages 273--290. CRC Press, 2008.

\bibitem{kong2020enhanced}
Gefei Kong and Hongchao Fan.
\newblock Enhanced facade parsing for street-level images using convolutional
  neural networks.
\newblock {\em IEEE Transactions on Geoscience and Remote Sensing},
  59(12):10519--10531, 2020.

\bibitem{kubanek2014capacities}
Julia Kubanek, Eike-Marie Nolte, Hannes Taubenb{\"o}ck, Friedemann Wenzel, and
  Martin Kappas.
\newblock Capacities of remote sensing for population estimation in urban
  areas.
\newblock {\em Earthquake Hazard Impact and Urban Planning}, pages 45--66,
  2014.

\bibitem{lee2013pseudo}
Dong-Hyun Lee et~al.
\newblock Pseudo-label: The simple and efficient semi-supervised learning
  method for deep neural networks.
\newblock In {\em Workshop on challenges in representation learning, ICML},
  volume~3, page 896, 2013.

\bibitem{li2020exploration}
Hao Li, Benjamin Herfort, Wei Huang, Mohammed Zia, and Alexander Zipf.
\newblock Exploration of openstreetmap missing built-up areas using twitter
  hierarchical clustering and deep learning in mozambique.
\newblock {\em ISPRS Journal of Photogrammetry and Remote Sensing}, 166:41--51,
  2020.

\bibitem{li2022improving}
Hao Li, Benjamin Herfort, Sven Lautenbach, Jiaoyan Chen, and Alexander Zipf.
\newblock Improving openstreetmap missing building detection using few-shot
  transfer learning in sub-saharan africa.
\newblock {\em Transactions in GIS}, 26(8):3125--3146, 2022.

\bibitem{liDevelopingMethodEstimate2020}
Xuecao Li, Yuyu Zhou, Peng Gong, Karen~C. Seto, and Nicholas Clinton.
\newblock Developing a method to estimate building height from {{Sentinel-1}}
  data.
\newblock {\em Remote Sensing of Environment}, 240:111705, April 2020.
\newblock \href {https://doi.org/10.1016/j.rse.2020.111705}
  {\path{doi:10.1016/j.rse.2020.111705}}.

\bibitem{liuIM2ELEVATIONBuildingHeight2020}
Chao-Jung Liu, Vladimir~A. Krylov, Paul Kane, Geraldine Kavanagh, and Rozenn
  Dahyot.
\newblock {{IM2ELEVATION}}: {{Building Height Estimation}} from {{Single-View
  Aerial Imagery}}.
\newblock {\em Remote Sensing}, 12(17):2719, January 2020.
\newblock \href {https://doi.org/10.3390/rs12172719}
  {\path{doi:10.3390/rs12172719}}.

\bibitem{milojevic2020learning}
Nikola Milojevic-Dupont, Nicolai Hans, Lynn~H Kaack, Marius Zumwald,
  Fran{\c{c}}ois Andrieux, Daniel de~Barros~Soares, Steffen Lohrey, Peter-Paul
  Pichler, and Felix Creutzig.
\newblock Learning from urban form to predict building heights.
\newblock {\em Plos one}, 15(12):e0242010, 2020.

\bibitem{pang20223d}
Hui~En Pang and Filip Biljecki.
\newblock 3d building reconstruction from single street view images using deep
  learning.
\newblock {\em International Journal of Applied Earth Observation and
  Geoinformation}, 112:102859, 2022.

\bibitem{parkCreating3DCity2019}
Yujin Park and Jean-Michel Guldmann.
\newblock Creating {{3D}} city models with building footprints and {{LIDAR}}
  point cloud classification: {{A}} machine learning approach.
\newblock {\em Computers, Environment and Urban Systems}, 75:76--89, May 2019.
\newblock \href {https://doi.org/10.1016/j.compenvurbsys.2019.01.004}
  {\path{doi:10.1016/j.compenvurbsys.2019.01.004}}.

\bibitem{raifer2019oshdb}
Martin Raifer, Rafael Troilo, Fabian Kowatsch, Michael Auer, Lukas Loos,
  Sabrina Marx, Katharina Przybill, Sascha Fendrich, Franz-Benjamin Mocnik, and
  Alexander Zipf.
\newblock Oshdb: a framework for spatio-temporal analysis of openstreetmap
  history data.
\newblock {\em Open Geospatial Data, Software and Standards}, 4:1--12, 2019.

\bibitem{redmon2018yolov3}
Joseph Redmon and Ali Farhadi.
\newblock Yolov3: An incremental improvement.
\newblock {\em arXiv preprint arXiv:1804.02767}, 2018.

\bibitem{resch2016impact}
Eirik Resch, Rolf~Andr{\'e} Bohne, Trond Kvamsdal, and Jardar Lohne.
\newblock Impact of urban density and building height on energy use in cities.
\newblock {\em Energy Procedia}, 96:800--814, 2016.

\bibitem{sunLargescaleBuildingHeight2019}
Yao Sun, Yuansheng Hua, Lichao Mou, and Xiao~Xiang Zhu.
\newblock Large-scale {{Building Height Estimation}} from {{Single VHR SAR}}
  image {{Using Fully Convolutional Network}} and {{GIS}} building footprints.
\newblock In {\em 2019 {{Joint Urban Remote Sensing Event}} ({{JURSE}})}, pages
  1--4, May 2019.
\newblock \href {https://doi.org/10.1109/JURSE.2019.8809037}
  {\path{doi:10.1109/JURSE.2019.8809037}}.

\bibitem{tost2019neural}
Heike Tost, Markus Reichert, Urs Braun, Iris Reinhard, Robin Peters, Sven
  Lautenbach, Andreas Hoell, Emanuel Schwarz, Ulrich Ebner-Priemer, Alexander
  Zipf, et~al.
\newblock Neural correlates of individual differences in affective benefit of
  real-life urban green space exposure.
\newblock {\em Nature neuroscience}, 22(9):1389--1393, 2019.

\bibitem{wang2020quiet}
Zhiyong Wang, Tessio Novack, Yingwei Yan, and Alexander Zipf.
\newblock Quiet route planning for pedestrians in traffic noise polluted
  environments.
\newblock {\em IEEE Transactions on Intelligent Transportation Systems},
  22(12):7573--7584, 2020.

\bibitem{wu2021roofpedia}
Abraham~Noah Wu and Filip Biljecki.
\newblock Roofpedia: Automatic mapping of green and solar roofs for an open
  roofscape registry and evaluation of urban sustainability.
\newblock {\em Landscape and Urban Planning}, 214:104167, 2021.

\bibitem{wurm2011object}
Michael Wurm, Hannes Taubenb{\"o}ck, Mathias Schardt, Thomas Esch, and Stefan
  Dech.
\newblock Object-based image information fusion using multisensor earth
  observation data over urban areas.
\newblock {\em International Journal of Image and Data Fusion}, 2(2):121--147,
  2011.

\bibitem{yan2019graph}
Xiongfeng Yan, Tinghua Ai, Min Yang, and Hongmei Yin.
\newblock A graph convolutional neural network for classification of building
  patterns using spatial vector data.
\newblock {\em ISPRS journal of photogrammetry and remote sensing},
  150:259--273, 2019.

\bibitem{yanEstimationBuildingHeight2022}
Yizhen Yan and Bo~Huang.
\newblock Estimation of building height using a single street view image via
  deep neural networks.
\newblock {\em ISPRS Journal of Photogrammetry and Remote Sensing}, 192:83--98,
  October 2022.
\newblock \href {https://doi.org/10.1016/j.isprsjprs.2022.08.006}
  {\path{doi:10.1016/j.isprsjprs.2022.08.006}}.

\bibitem{yuan_knowledge_2022}
Zhendong Yuan, Jules Kerckhoffs, Gerard Hoek, and Roel Vermeulen.
\newblock A knowledge transfer approach to map long-term concentrations of
  hyperlocal air pollution from short-term mobile measurements.
\newblock {\em Environmental Science \& Technology}, September 2022.
\newblock \href {https://doi.org/10.1021/acs.est.2c05036}
  {\path{doi:10.1021/acs.est.2c05036}}.

\bibitem{zhang2021vgi3d}
Chaoquan Zhang, Hongchao Fan, and Gefei Kong.
\newblock Vgi3d: an interactive and low-cost solution for 3d building modelling
  from street-level vgi images.
\newblock {\em Journal of Geovisualization and Spatial Analysis}, 5(2):1--16,
  2021.

\bibitem{zhangBuildingHeightExtraction2022}
Chenni Zhang, Yunfan Cui, Zeyao Zhu, San Jiang, and Wanshou Jiang.
\newblock Building {{Height Extraction}} from {{GF-7 Satellite Images Based}}
  on {{Roof Contour Constrained Stereo Matching}}.
\newblock {\em Remote Sensing}, 14(7):1566, January 2022.
\newblock \href {https://doi.org/10.3390/rs14071566}
  {\path{doi:10.3390/rs14071566}}.

\bibitem{zhaoCBHECornerbasedBuilding2019}
Yunxiang Zhao, Jianzhong Qi, and Rui Zhang.
\newblock {{CBHE}}: {{Corner-based Building Height Estimation}} for {{Complex
  Street Scene Images}}.
\newblock In {\em The {{World Wide Web Conference}}}, {{WWW}} '19, pages
  2436--2447, {New York, NY, USA}, May 2019. {Association for Computing
  Machinery}.
\newblock \href {https://doi.org/10.1145/3308558.3313394}
  {\path{doi:10.1145/3308558.3313394}}.

\bibitem{zhu2005semi}
Xiaojin~Jerry Zhu.
\newblock {\em Semi-supervised learning literature survey}.
\newblock University of Wisconsin-Madison Department of Computer Sciences,
  2005.

\end{thebibliography}

\end{document}